\documentclass[onefignum, onetabnum,onealgnum, onethmnum]{siamonline190516}

\usepackage{comment}
\usepackage{lipsum}
\usepackage[margin=1in, footskip=36pt]{geometry}

\usepackage{endnotes}

\usepackage{titlesec}
\usepackage{titletoc}
\usepackage{pgfplotstable}
\titleclass{\task}{straight}[\section]
\newcounter{task}
\renewcommand{\thetask}{\arabic{task}}
\titleformat{\task}[hang]
    {\normalfont\LARGE\bfseries}{Task \thetask:}{1em}{}

\titleformat*{\task}{\color{header1}\bfseries}

\titlecontents{task}
              [3.8em] 
              {}
              {\contentslabel{2.3em}}
              {\hspace*{-2.3em}}
              {\titlerule*[1pc]{.}\contentspage}

\titlespacing*{\section}{0ex}{1ex}{1ex}
\titlespacing*{\subsection}{0ex}{1ex}{1ex}
\titlespacing*{\subsubsection}{0ex}{1ex}{1ex}
\titlespacing*{\paragraph}{0ex}{1ex}{1ex}
\titlespacing*{\subparagraph}{0pt}{1ex}{1ex}
\titlespacing*{\task}{0em}{1ex}{1ex}

\newcommand\blfootnote[1]{%
  \begingroup
  \renewcommand\thefootnote{}\footnote{#1}%
  \addtocounter{footnote}{-1}%
  \endgroup
}

\usepackage[sort&compress,comma,square,numbers]{natbib}

\usepackage{amsfonts,amsmath,amssymb}
\usepackage{bbm}

\usepackage{multicol}
\usepackage{enumitem}
\setlist[enumerate]{wide, labelindent=1cm,  noitemsep}
\setlist[itemize]{noitemsep}
\setlist[description]{noitemsep}

\usepackage{hhline}

\RequirePackage{ulem}

\definecolor{tableheadcolor}{gray}{0.92}
\newcommand{\topline}{ %
        \arrayrulecolor{blue1}\specialrule{0.1em}{\abovetopsep}{0pt}%
        \arrayrulecolor{tableheadcolor}\specialrule{\belowrulesep}{0pt}{0pt}%
        \arrayrulecolor{blue1}}
\newcommand{\midtopline}{ %
        \arrayrulecolor{tableheadcolor}\specialrule{\aboverulesep}{0pt}{0pt}%
        \arrayrulecolor{blue1}\specialrule{\lightrulewidth}{0pt}{0pt}%
        \arrayrulecolor{white}\specialrule{\belowrulesep}{0pt}{0pt}%
        \arrayrulecolor{blue1}}
\newcommand{\bottomline}{ %
        \arrayrulecolor{white}\specialrule{\aboverulesep}{0pt}{0pt}%
        \arrayrulecolor{blue1} %
        \specialrule{\heavyrulewidth}{0pt}{\belowbottomsep}}%

\pgfplotstableset{normal/.style ={%
        header=true,
        string type,
        font=\small,
    	column type={p{.092\textwidth}},
        every head row/.style={
            before row={\topline\rowcolor{tableheadcolor}},
            after row={\midtopline}
        },
        every odd row/.style={
            before row={\rowcolor{blue1}}
        },
        every last row/.style={
            after row=\bottomline
        },
        col sep=&,
        row sep=\midrule
    }
}

\usepackage{multirow}

\usepackage{booktabs}
\usepackage{titlesec}
\usepackage{fancyhdr}
\usepackage{fullpage}

\RequirePackage{titlesec}
\RequirePackage[font={footnotesize},{singlelinecheck=false}]{caption}
\usepackage[T1]{fontenc}
\usepackage[scaled]{helvet}
\usepackage[scaled=1.10, med]{zlmtt}

\usepackage{algorithm}
\usepackage{algpseudocode}

\newcommand{\T}{^{\ensuremath{\mathsf{T}}}} 

\newcommand{\Real}{\mathbb{R}}

\title{
When are Deep Networks really better than Decision Forests at small sample sizes, and how?
}

\author{%
  Haoyin Xu$^{1,*}$,
  Kaleab A. Kinfu$^{1}$,
  Will LeVine$^{1}$,
  Sambit Panda$^{1}$,
  Jayanta Dey$^{1}$,
  Michael Ainsworth$^{1}$,
  Yu-Chung Peng$^{1}$,
  Madi Kusmanov$^{1}$,
  Florian Engert$^{2}$,
  Christopher M. White$^{3}$,
  Joshua T. Vogelstein$^{1}$, and
  Carey E. Priebe$^{1}$
}

\pgfplotsset{compat=1.15}
\begin{document}
\maketitle
\blfootnote{
  $^1$Johns Hopkins University
  $^2$Harvard University
  $^3$Microsoft Research
  $^*$Corresponding author:
  \href{mailto:hxu36@jhu.edu}{hxu36@jhu.edu}
  
  The authors acknowledge the National Science Foundation-Simons Foundation’s Research Collaboration on the Mathematical and Scientific Foundations of Deep Learning (MoDL), NSF grant 2031985.
}
\pagenumbering{arabic}
\setcounter{page}{1}

\begin{abstract}
Deep networks and decision forests (such as random forests and gradient boosted trees) are the leading machine learning methods for structured and tabular data, respectively.
Many papers have empirically compared large numbers of classifiers on one or two different domains (e.g., on 100 different tabular data settings). 
However, a careful conceptual and empirical comparison of these two strategies using the most contemporary best practices has yet to be performed.
Conceptually, we illustrate that both can be profitably viewed as ``partition and vote'' schemes. Specifically, the representation space that they both learn is a \textit{partitioning} of feature space into a union of convex polytopes. For inference, each decides on the basis of \textit{votes} from the activated nodes. This formulation allows for a unified basic understanding of the relationship between these methods. 
Empirically, we compare these two strategies on hundreds of tabular data settings, as well as several vision and auditory settings. 
Our focus is on datasets with at most 10,000 samples, which represent a large fraction of scientific and biomedical datasets.
In general, we found forests to excel at tabular and structured data (vision and audition) with small sample sizes, whereas deep nets performed better on structured data with larger sample sizes.
This suggests that further gains in both scenarios may be realized via further combining aspects of forests and networks.
We will continue revising this technical report in the coming months with updated results.
\end{abstract}

\section{Introduction}
In the last decade, decision forests (forests hereafter) and deep networks (networks hereafter) have gained prominence in the scientific literature as two of the highest performing techniques for machine learning tasks, including classification. Forests have empirically dominated \textit{tabular} data scenarios, where the relative position of features is irrelevant. More specifically, random forests (RF) and gradient boosted trees (GBDT) have outperformed all other methods in papers comparing various methods on real datasets and machine learning competitions, demonstrating their relevance to the many biomedical applications represented by tabular data \citep{gbdt, Caruana2006-wp, Caruana2008-tb, JMLR:v15:delgado14a, Chen2016-fx}. In contrast, networks typically dominate other methods on large sample size \textit{structured} data scenarios, where the relative position of features is key for sample identification. Those include vision, audition, text, and autonomous control \citep{Krizhevsky2012-sq, Zhang2020-vg, Brown2020-tz, Statt2019-ox, resnet}.

Nonetheless, the fact that these two approaches dominate in complementary settings have motivated a number of efforts to combine the best of both worlds. 
For example, \citet{Patel2015-jg} pointed out that under certain assumptions, both forests and networks can be cast as max-sum message passing networks, \citet{Zhou2018-ii} combined deep networks with random forests, \citet{Shen2019-mq} has worked to make random forests differentiable, and others have made random forests end-to-end trainable \citep{Carreira-Perpinan2018-py, Hehn2019-kh}. However, the relationship between the internal representations that the two approaches learn has not yet been made explicit, to our knowledge. Here, we illustrate the conceptual commonalities of their representations \citep{Priebe2020.04.29.068460}.

The ``arbitrarily slow convergence'' and the ``no free lunch'' theorems prove that no one approach can outperform another approach in all problems \citep{slow_conv, lunch}. While forests and networks are commonly studied to determine their optimal usage, we also compare them empirically to establish guidelines informing the community in which each approach outperforms the other. We do so using tabular, vision, and auditory data, varying training set sample sizes ranging from only a few samples per class up to 10,000 total for each dataset.
We limit the sample size because small sample size problems remain a huge challenge in data science, and are essentially ubiquitous in scientific and biomedical datasets.

Viewed conceptually as ``partition and vote'' schemes, both forests and networks partition feature space into a union of convex polytopes and predict by voting from the activated nodes. The similarities provide a unified basic understanding of the relationship between these methods and could be leveraged to explore learning in biological brains. 
Empirically, forests excel at tabular and structured data (vision and audition) with small sample sizes, whereas networks perform better on structured data with larger sample sizes. This suggests that further gains in both scenarios may be realized via further combining aspects of forests and networks.



\section{Conceptual Similarities}
The classical statistical formulation of the classification problem consists of $(X,Y), (X_1,Y_1), \cdots, (X_n,Y_n) \stackrel{iid}{\sim} F_{XY}$, where $\mathcal{T}_n = \{(X_i,Y_i)\}_{i \in \{1,\cdots,n\}}$ is the training data and $(X,Y)$ represents the to-be-classified test observation $X$ with true-but-unobserved class label $Y$. We consider the simplest setting in which $X$ is a feature vector in $\Real^d$ and $Y$ is a class label in $\{0,1\}$. The goal is to learn a classification rule $g_n = g(\cdot;\mathcal{T}_n)$ mapping feature vectors to class labels such that the probability of misclassification $L(g_n) = P[g(X;\mathcal{T}_n) \neq Y|\mathcal{T}_n]$ is small.

Stone's theorem for universally consistent classification \citep{Stone1977, DGL} demonstrates, loosely speaking, that a successful classifier can be constructed by first partitioning the input space into cells depending on $n$, such that the number of training data points in each cell goes to infinity but slowly in $n$, and then estimating the posterior 
$\eta(x) = P[Y=1|X=x]$ locally by voting based on the training class labels associated with the training feature vectors in cell $C(x) \subset \Real^d$ in which the test observation falls.
Then, under some technical conditions $L(g_n) \to L^*$ almost surely for any $F_{XY}$, where $L^*$ is the Bayes optimal probability of misclassification.

In the context of our formulation of the classification problem,
we provide a unified description of the two dominant methods in modern machine learning, forests and networks, as ensemble ``partition and vote'' schemes.
This allows for useful basic insight into their relationship with each other, and potentially with brain functioning.

\subsection{Decision Forests}
Forests, including RF and GBDT, demonstrate state-of-the-art performance in a variety of machine learning settings.
Forests have typically been implemented as ensembles of axis-aligned decision trees, but extensions also employ axis-oblique splits \citep{Amit1997, Breiman2001, Tomita2017-mv, sporf}.

\begin{figure}[htb]
\centering
\includegraphics[width=0.8\textwidth]{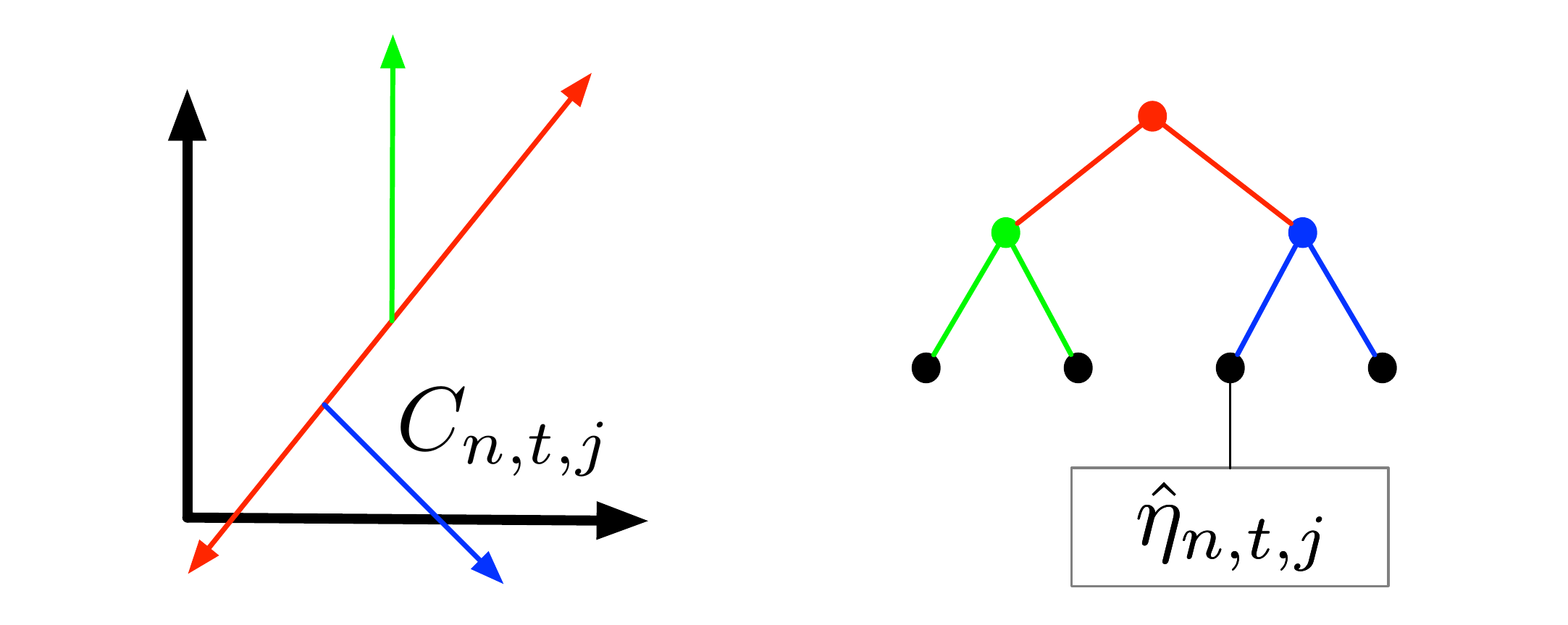}
  \caption{A tree in the forest.
  Given the random subset $\mathcal{T}_{n,t}$ of the training data allocated to tree $t$, the root node (red) performs a hyperplane split of $\Real^d$ based on a random subset of dimensions;
  the two daughter nodes (green and blue) split their respective partition cells based on a separate random subset of dimensions allocated to each node; etc.
  In the end, this tree results in a partition of $\Real^d$ with the leaf nodes corresponding to partition cells for which training data class labels yield local posterior estimates.
  The forest classifies the test observation feature vector $X$ by voting over trees using the cells $C_{n,t}(X)$ in which $X$ falls.
  }
\label{fig:forest}
\end{figure}

Given the training data $\mathcal{T}_n$, each tree $t$ in a RF constructs a partition $\mathcal{P}_{n,t}$ by successively splitting the input space based on a random subset of data (per tree) and then choosing a hyperplane split (Figure \ref{fig:forest}) based on a random subset of dimensions (per node) using some criterion for split utility at each node. The depth of each tree is a function of $n$ and involves a tradeoff between leaf purity and regularization to alleviate overfitting.
Details aside, each tree results in a partition $\mathcal{P}_{n,t}$ of $\Real^d$
and each partition cell---each leaf of each tree---admits a posterior estimate $\widehat{\eta}_{n,t,j} = (1 / N_{n,t,j}) \sum_{i: X_i \in C_{n,t,j}} I\{Y_i=1\}$ based on the class labels of the training data feature vectors in cell $C_{n,t,j}$.
Under appropriate conditions for partitions $\mathcal{P}_{n,t}$ we have $L(g^{RF}_n) \to L^*$.

\subsection{Deep Networks}
\begin{figure}[htb]
\centering
\includegraphics[width=0.8\textwidth]{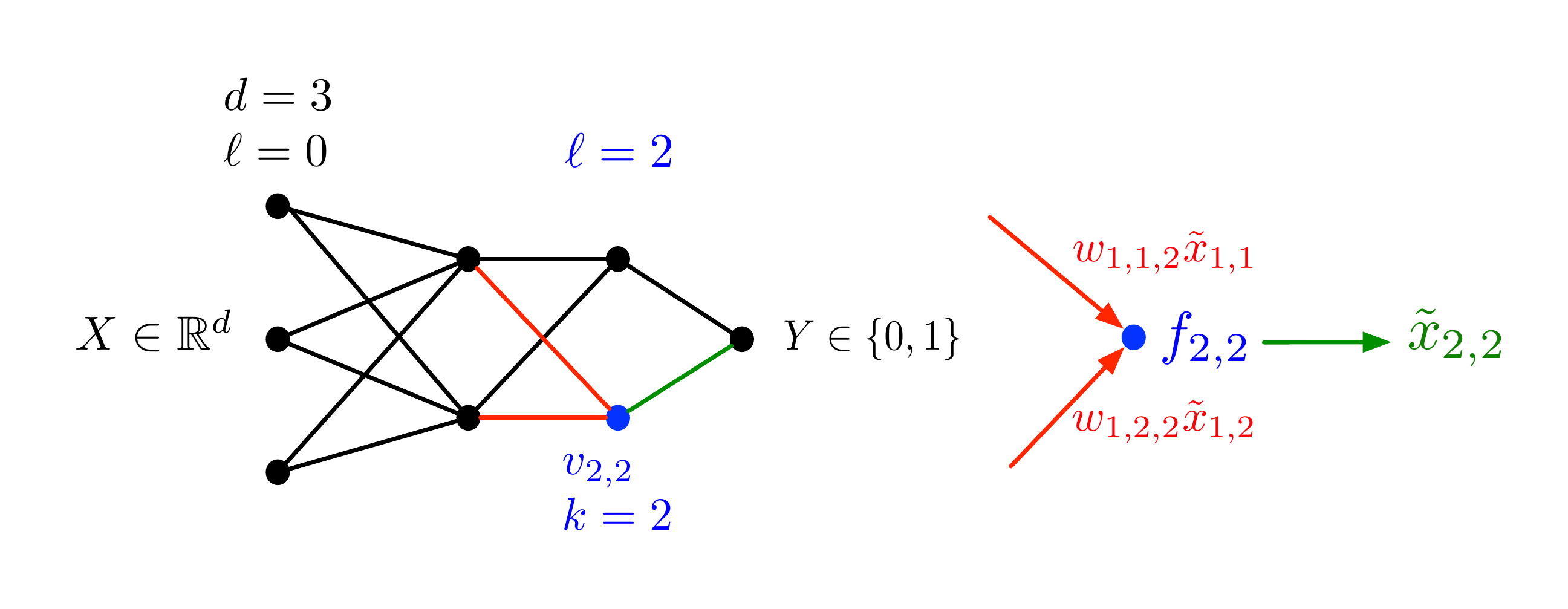}
  \caption[]{A (shallow) deep network.
  Given the training data $\mathcal{T}_{n}$, the $X_i \in \Real^d$ are passed through the network.
  At layer $\ell$ and node $v_{\ell,k}$ (the blue node is $v_{2,2}$) the inputs $w_{\ell-1,j,k} \tilde{x}_{\ell-1,j}$ (red) are transformed via hyperplane activation function $f_{\ell,k}$ and output as $\tilde{x}_{\ell,k}$ (green).
  Thus node $v_{2,2}$ receives non-zero input $w_{1,j,2} \tilde{x}_{1,j} = w_{1,j,2} (X\T W_{0,j} + b_{1,j})^+$ from node $v_{1,j}$ if and only if the linear combination of the multivariate $X$, $X\T W_{0,j}$, is on the preferred side of hyperplane defined by $f_{1,j}$ (and weight $w_{1,j,2}$ is non-zero).
  The output of node $v_{2,2}$ is 
  $\tilde{x}_{2,2} = \left( \left[\substack{(X\T W_{0,1} + b_{1,1})^+\\(X\T W_{0,2} + b_{1,2})^+}\right]\T W_{1,2} + b_{2,2} \right)^+$; 
  that is, $v_{2,2}$ provides a further hyperplane refinement of $\Real^d$.
  }
\label{fig:network}
\end{figure}

Networks are extraordinarily popular and successful in modern machine learning \citep{lecun2015deeplearning, SzeChe17Efficient, MPCB2014, M2017}.
Given a network of nodes and edges (Figure \ref{fig:network}), each internal node $v_{\ell,k}$ in layer $\ell$ of the network gathers inputs $\tilde{x}_{\ell-1,j}, j=1,\cdots,n_{\ell-1}$ from the previous layer, weighted by $w_{\ell-1,j,k}, j=1,\cdots,n_{\ell-1}$, and outputs 
$
\tilde{x}_{\ell,k} 
= f_{\ell,k}(\sum_j \tilde{x}_{\ell-1,j} w_{\ell-1,j,k})
= f_{\ell,k}(\tilde{X}_{\ell-1}\T W_{\ell-1,k})
= f(\tilde{X}_{\ell-1}\T W_{\ell-1,k} + b_{\ell,k}).
$
When using the ReLU (rectified linear unit) function $\max(0,\cdot) = (\cdot)^+$ as the activation function $f$, each node performs a hyperplane split based on a linear combination of its inputs, passing a non-zero value forward if the input is on the preferred side of the hyperplane; data in the cell defined by the collection of polytopes induced by nodes $\{v_{\ell-1,j}\}$, weighted, falls into node $v_{\ell,k}$
and is output based on a partition refinement.
Thus, a node in the last internal layer corresponds to a union of hyperplane-induced partition cells, defined via composition of all earlier nodes in the network.

To create a conceptual unification with forests, consider passing all the training data $X$ through the network; for a network, the input $X$ falls into a final network partition cell in the last internal layer \citep{MPCB2014}.
This partition cell is encoded by the set of nodes in the penultimate layer that are activated by $X$.
This approach differs from forests: in a forest, the ensemble is realized by voting over a forest of trees, while in a network each $X$ can activate many (even all) cells in the penultimate layer, though with different activation energies.
In other words, both forests and networks can be seen to use the same representation space, 
though they achieve their particular representation via different estimation (``learning'') algorithms. Specifically, they both learn piecewise linear activation functions \citep{Serra2018-bg, shi2019gradient, Rolnick2019-ei, Hanin2019-jx}.

\subsection{A Union of Convex Polytopes}
\begin{figure}[htb]
\centering
\includegraphics[width=0.3\textwidth]{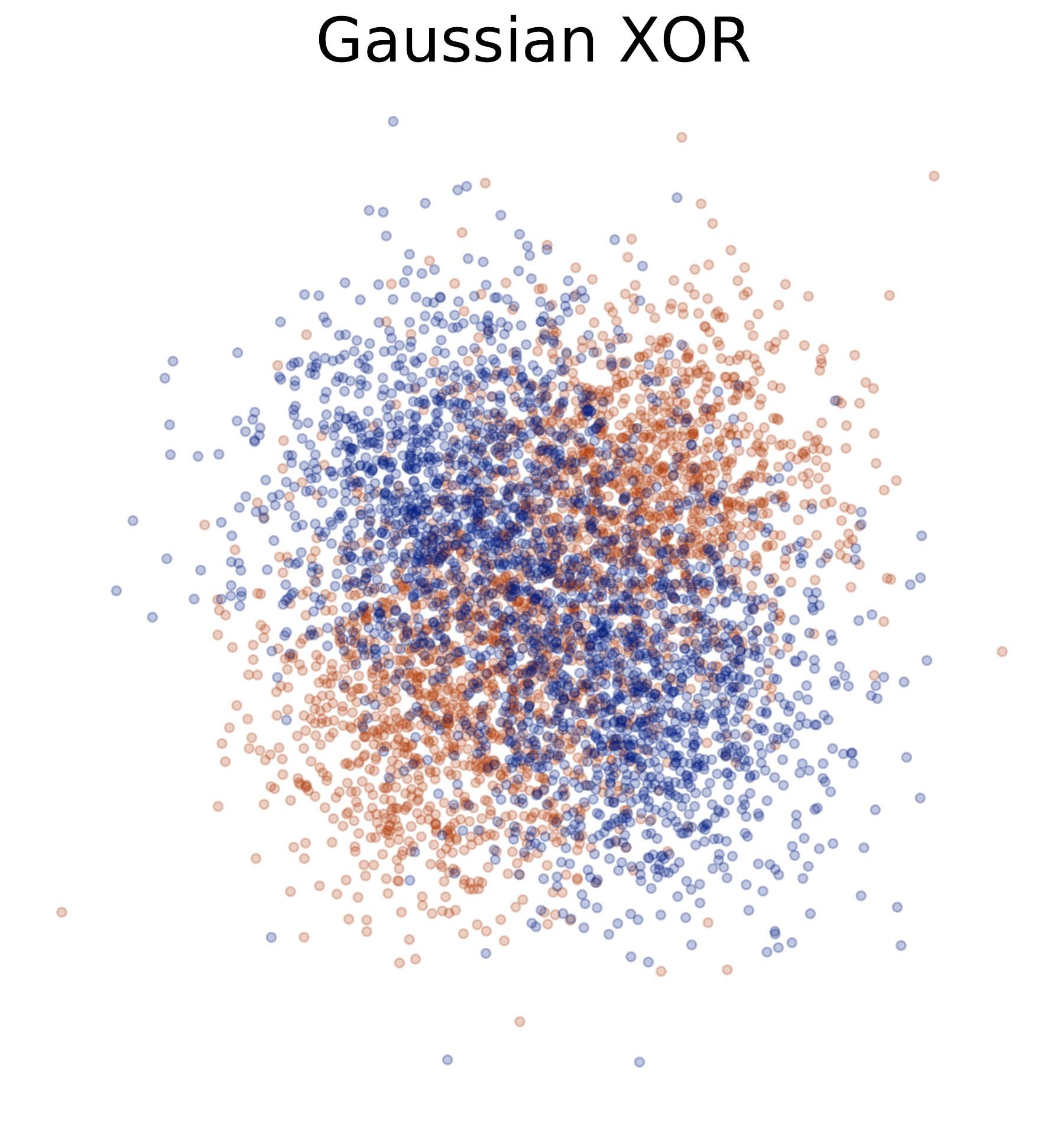}
\includegraphics[width=0.65\textwidth]{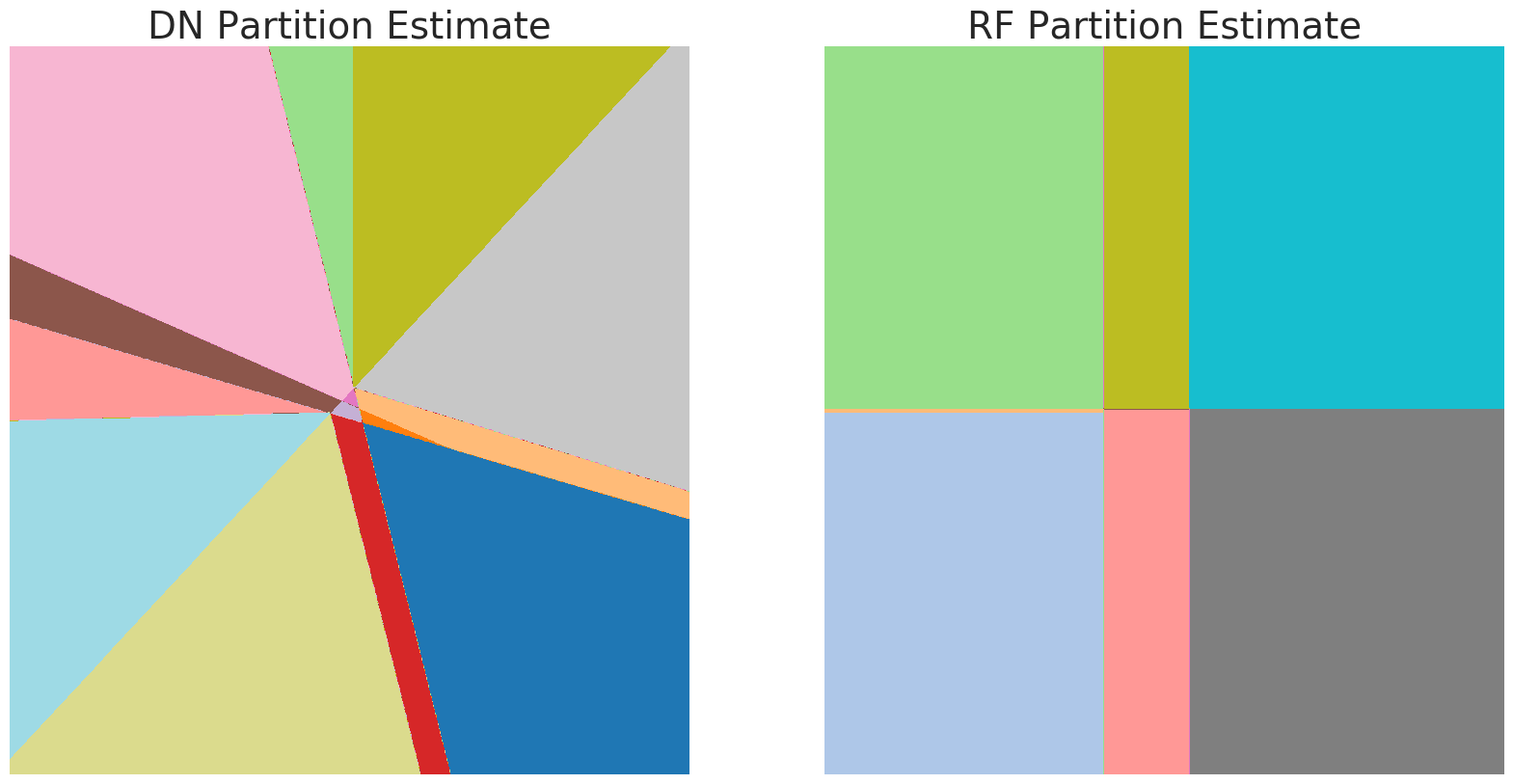}
  \caption{\textbf{(Left)} Gaussian XOR samples used as a benchmark. 
  \textbf{(Center and right)} Visualizations of the partitions defined by $C_L$ with respect to input space, $\mathcal{X}$, defined by a network \textbf{(center)} and a forest \textbf{(right)}. 
  A unique region (i.e. part) is visualized by an arbitrary unique color and corresponds to a specific activation region.
  Note that the color value does not have any particular meaning. 
  }
\label{fig:xor}
\end{figure}

To provide a concrete illustrative example, consider the following experimental setup. We generated a two-dimensional Gaussian XOR dataset (Figure \ref{fig:xor}, left) as a benchmark with four spherically symmetric Gaussians. Class 0 has two Gaussians with centers $(-1,-1)$ and $(1,1)$, whereas class 1 has two Gaussians with centers $(1,-1)$ and $(-1,1)$. There are 5,096 random samples from the two classes, which are split into 4,096 training samples and 1,000 test samples. We trained both forests and networks on such data. Figure \ref{fig:xor} (center and right) shows the partitions learned by the two methods. More formally, let $N_l$ be the set of nodes in a given layer $l$ and let $A_l(x) = \{a \in N_l | a(x) > 0\}$, where $a$ is the activation function of the underlying forests/networks. That is, $A_l$ takes in an inference example as input and outputs the set of nodes \textit{in a given layer} on which the input inference example activates. Now let $A_L = \bigoplus_{l = 0}^L A_l$.
Given an inference example as input, $A_L$ outputs the concatenated set of nodes \textit{in all layers up to and including layer $L$} on which the input inference example activates. 
In Figure \ref{fig:xor} (center and right), similarly colored points output the same value of $A_L$. 
These collections of points are partitions, defined by $C_L(x) = \{z \in \mathcal{X} | A_L(z) = A_L(x)\}$ for a given inference point $x$, where $\mathcal{X}$ is the domain of the forests/networks.
In forests with $L$ equal to the depth, this partition is a leaf cell. In networks with $L$ equal to the total number of layers, this partition is a convex polytope.

\begin{figure}
\centering
\includegraphics[width=1.0\textwidth]{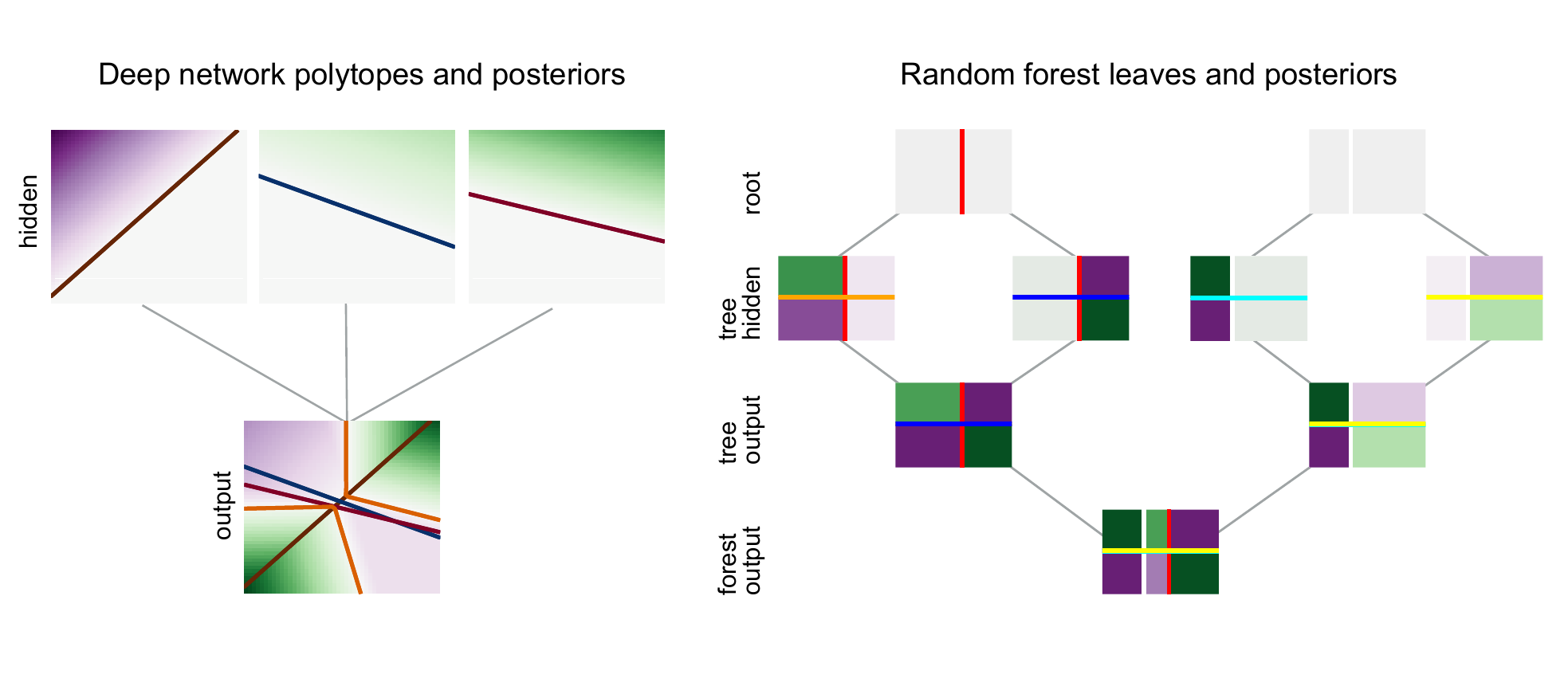}
  \caption{Visualizations of the polytope compositions for networks \textbf{(left)} and forests \textbf{(right)}. For the first layer/depth polytopes/leaves, we show the boundary for which $C_0$ changes value. That is, on one side of the visualized boundary the node is activated; on the other side, the node is not activated. In the case of networks, this means that one side contains the linear portion of the ReLU activation, while the other side contains the 0-valued region of the ReLU activation; in the case of forests, this indicates to which subsequent node (either left or right) examples will recursively fall. For all polytopes/leaves in all subsequent layers/depths, we visualize the $C_l$ boundaries for that layer, and we also overlay the boundaries of $C_{l`}$ for all previous layers $l` < l$. The bottom rows both indicate the final model cells $C_L$. 
  In all nodes, the magnitude of the background color is determined by the activations of that layer, with class 0 being more purple, and class 1 being more green. 
  }
\label{fig:posteriors}
\end{figure}

In Figure \ref{fig:posteriors}, we visualize the layer-wise composition of $C_L$, the boundaries of $C_l$ for all layers and for each node in each layer, and the effect of $C_l$ on posteriors. This provides additional context for the creation of Figure \ref{fig:xor}.

\section{Empirical Differences}
\subsection{Tabular}
\label{tabular}
We used OpenML-CC18 for tabular benchmarks, which is a collection of 72 datasets organized by OpenML to function as a comprehensive benchmark suite \citep{OpenML2013, bischl}. These datasets vary in sample size, feature space, and unique target classes.
About half of the datasets are binary classification tasks, and the other half are classification tasks with up to 50 classes. The range of total sample sizes of the datasets is between 500 and 100,000, while the range of features is from a few to a few thousand.
Datasets were imported using the OpenML python package, which is BSD-licensed \citep{OpenMLPython2019}.

\subsubsection{Methods}
\paragraph{Computing}
All datasets with over 10,000 samples were randomly downsampled to 10,000 samples. Then benchmarks were performed using 5-fold cross-validation, with the held-out test folds used to evaluate all classification tasks for the given dataset.
Next, for each dataset's training folds, the training data were indexed into eight subsets with evenly spaced sample sizes on a logarithmic scale, thus producing eight training sets with different sample sizes. The smallest of these sample sizes had five times the number of classes in the dataset, and the largest sample size used the entire training fold.

Random forest is a non-parametric and universally consistent estimator, so it will approach Bayes optimal performance with sufficiently large sample sizes, tree depths, and trees \citep{biau}.
We used 500 trees with no depth limit, only varying the number of features selected per split (``max-features'').
Letting $d$ equal the number of features in the dataset, we varied max-features to be one of: $\sqrt{d}$, $d/4$, $d/3$, $d/1.5$, and $d$ \citep{Probst2019}.

For optimizing hyperparameters in a network, we essentially followed the guidance of \citet{bouthillier}. \texttt{MLPClassifier} from the scikit-learn (BSD 3-Clause) package was used with one hidden layer \citep{scikit-learn}. The default setting was used for weight initialization, and the following parameters were tuned: hidden layer size between 20 and 400, and the L2 regularization parameter along log-uniform from $1 \times 10^{-5}$ to $1 \times 10^{-2}$. These parameters were determined based on previous classification work by \citet{jurtz} on the amino acid dataset \citep{MHC}.
To extend this optimization to over-parametrized networks, we also searched over the number of layers of the network from one to three hidden layers spanning all combinations of node sizes from 20 to 400. 

For all hyperparameter tuning in both forests and networks, the tuning was conducted only on the entire dataset, using a randomized hyperparameter search with five folds. Tuned hyperparameters were then consistent for all smaller sample sizes per dataset. All tabular benchmarks were run on a 2.3 GHz 8-core Intel i9 CPU. Model training and hyperparameter tuning were parallelized using all available cores.

\paragraph{Evaluation Criteria}
Cohen's Kappa ($\kappa = \frac{p_o - p_c}{1 - p_c}$, where $p_o$ is the proportion of agreements, and $p_c$ is the expected proportion of chance agreements) is an evaluation metric between two raters that determines the level of agreement between each \citep{cohen}.
Unlike classification accuracy, Cohen's Kappa normalizes its score by accounting for chance accuracy. It can be a powerful tool for this experiment because, at small sample sizes, chance accuracy may have a large impact on the model evaluation. In the case of supervised classification, the two raters represent the predictions of the machine learning models and the ground truth of the data. The mean Kappa value was then recorded across the five folds for every sample size. A higher number represents higher accuracy, where a perfectly accurate model has a Kappa value of one.

Expected calibration error (ECE) is a metric used to compare two distributions by calculating the expected difference between accuracy and confidence \citep{naeini2015obtaining}. In addition to Cohen's Kappa, ECE was computed for each dataset at each sample size. 
This method is executed by storing predictions in $N = 40$ equally spaced bins and calculating the weighted averages of the bins' differences in accuracy versus confidence \citep{pmlr-v70-guo17a}.
A lower number represents higher calibration, where a perfectly calibrated model has an ECE of zero.

Lastly, training wall times were recorded for both models. This metric calculated the fitting time for the given model after hyperparameter tuning, measured in seconds.

\subsubsection{Results}

\begin{figure}[htb]
\centering
\includegraphics[width=1.0\textwidth]{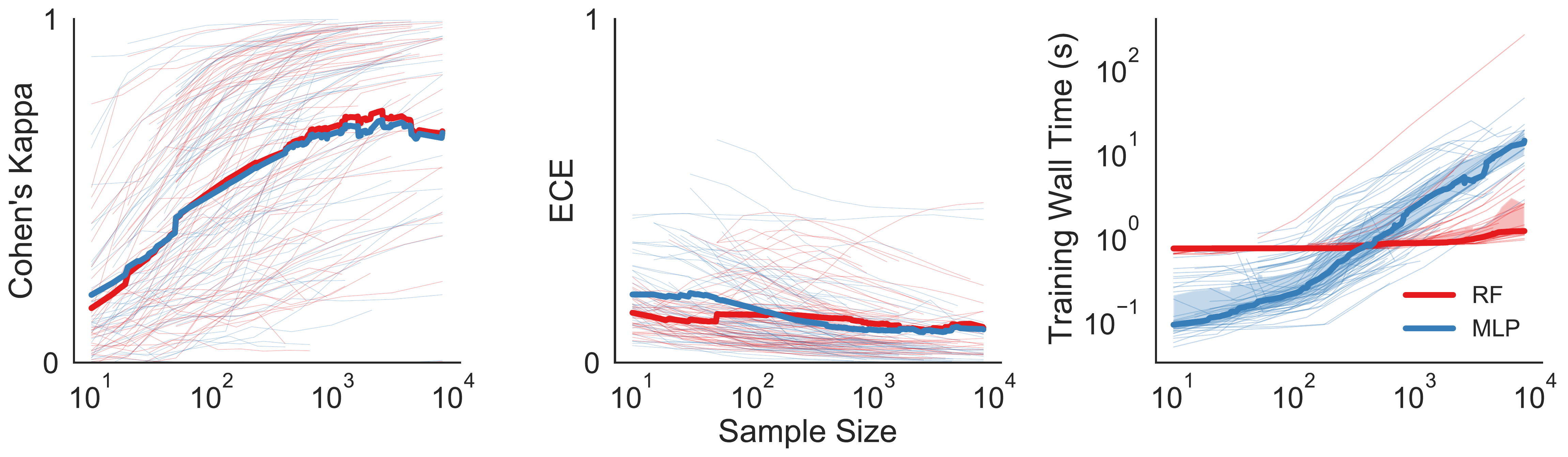}
  \caption{Performance of forests and networks, evaluated using Cohen's Kappa \textbf{(left)}, expected calibration error (ECE) \textbf{(center)}, and training wall times \textbf{(right)}. 
  For all plots, thin lines represent individual datasets, and shaded regions highlight the 25th through 75th percentiles. Left plot shows Cohen's Kappa versus sample size. Higher values mean better performance. Center plot shows ECE versus sample size. Lower values mean better calibration. Thick lines in both plots are the interpolated means. Right plot shows training wall times versus sample size. Thick lines are the interpolated medians. 
  At larger sample sizes, Cohen's Kappa shows that forests are more accurate, while ECE suggests that networks are better calibrated. For training wall times, networks are faster than forests at small sample sizes but slower at larger sample sizes.
  }
\label{fig:tab}
\end{figure}

\paragraph{Cohen's Kappa and ECE}
Sample sizes are plotted on a logarithmic scale, whereas Cohen's Kappa and ECE performance are plotted linearly (Figure \ref{fig:tab}). As seen from these results, there is a high level of variability between tabular datasets at each sample size. At larger sample sizes, forests tend to win Cohen's Kappa and excel at accuracy, but networks would win ECE and achieve better calibration.

\paragraph{Training Wall Times}
Sample size have little impact on forest training wall times, yet there exists an overhead cost of slightly less than one second to fit these models (Figure \ref{fig:tab}). Networks, on the other hand, are an order of magnitude faster than forests at small sample sizes, but quickly increase to be an order of magnitude slower as sample size increase.

Given the OpenML-CC18 tabular dataset suite, there appears to be little relationship between average model performance and sample size. However, tabular benchmarks (Figure \ref{fig:tab}) show trends differentiating the classifiers' ECE scores, which suggest that forests could produce better-calibrated models at small sample sizes. Should model calibration be the priority, this difference yields a framework for selecting the optimal method given the available sample size on general tabular data. The behavior of training wall times also suggests a trade-off for method choices. At larger sample sizes, network training times would scale much faster. If time is a limiting factor, then forests may be the ideal selection for novel tabular data at larger sample sizes.


\subsection{Vision}
\label{vision}
We used CIFAR-10 and CIFAR-100 datasets to evaluate the performance of forests and networks on vision data, primarily because of the number of classes and large sample sizes of these datasets \citep{cifar}.
Each dataset contains 60,000 colored images with 32x32 pixels, which are separated into 10 or 100 classes, resulting in 6,000 or 600 images per class.
We also included the SVHN dataset as a supplement (Appendix \ref{app:svhn}) \citep{svhn}. 
Each high-dimensional image sample would be represented by RGB pixels and thus contain $32 \times 32 \times 3 = 3,072$ features.

\subsubsection{Methods}
\paragraph{Computing}
We experimented with multi-class classifications in 3-class, 8-class, and 90-class settings. We sampled the 3-class and 8-class training sets from the CIFAR-10 and SVHN datasets, and the 90-class training sets from the CIFAR-100 dataset. For each classification task, we ensured 45 random combinations of class labels and up to 10,000 training samples, stratifying data equally across the classes.

We used \texttt{RandomForestClassifier} from the scikit-learn python package \citep{scikit-learn}. For the deep learning models, four convolutional neural network (CNN) architectures using ReLU activation were employed: three simpler models built with varying parameters and ResNet-18 \citep{Priebe2020.04.29.068460, resnet}. We adapted the pre-trained ResNet-18 classifier from PyTorch (BSD-3) as a robust choice and optimized its last layer with our training sets \citep{pytorch}. Among the three simpler CNNs, the 1-layer CNN consists of one convolutional layer of 32 filters and one fully connected layer. The 2-layer CNN consists of two convolutional layers each with 32 filters, followed by two fully connected layers with 100 and 10 nodes, respectively. Lastly, the 5-layer CNN scales up to 128 feature maps, and each layer is followed by batch normalization.

For the high-dimensional vision data, we minimized tuning and benchmarked models with their default values ``out of the box'' \citep{Probst2019hy}. For RF, we used scikit-learn's default parameters for all model aspects except parallelizing with all the cores \citep{scikit-learn}. We benchmarked forests on two Microsoft Azure compute instances: a 2-core Standard\_DS2\_v2 (Intel Xeon E5-2673 v3) and a 6-core Standard\_NC6 (Intel Xeon E5-2690 v3) (Table \ref{table:azure}).

\begin{table}[htb]
\centering
\begin{tabular}{ |c|c|c|c|c|c| } 
\hline
Compute Instance & vCPU & Memory: GiB & SSD: GiB & GPU & GPU Memory: GiB \\
\hline
Standard\_DS2\_v2 & 2 & 7 & 14 & 0 & N/A \\
\hline
Standard\_NC6 & 6 & 56 & 340 & 1 & 12 \\
\hline
\end{tabular}
\caption{Specifications for Azure compute instances.}
\label{table:azure}
\end{table}

We implemented all network models with a learning rate of 0.001 and a batch size of 64, using stochastic gradient descent with a momentum of 0.9 and cross-entropy loss. These hyperparameters were chosen either by default or commonly seen in literature, and for training iterations, our settings provided the model with enough time to visually converge on the loss \citep{Krizhevsky2012-sq, pmlr-v119-rice20a}.
We first implemented the networks with 30 epochs and activated early stopping when validation loss did not improve in three epochs in a row \citep{li2020, lutz, caruana}. In these tasks, we randomly selected 30\% of the provided test sets as validation data and only used the held-out 70\% for benchmarks.

Alternative approaches restricted CNN epochs by calibrating their training wall times or money cost, conforming to those of RF as run on different compute instances. The alternative approaches used the full data of the provided test sets for benchmarks. All approaches were benchmarked on the 1-core GPU component of a Microsoft Azure compute instance: Standard\_NC6 (NVIDIA Tesla K80).
We utilized the GPU with the PyTorch CUDA library \citep{pytorch}.

\begin{figure}[htb]
\centering
\includegraphics[width=0.8\textwidth]{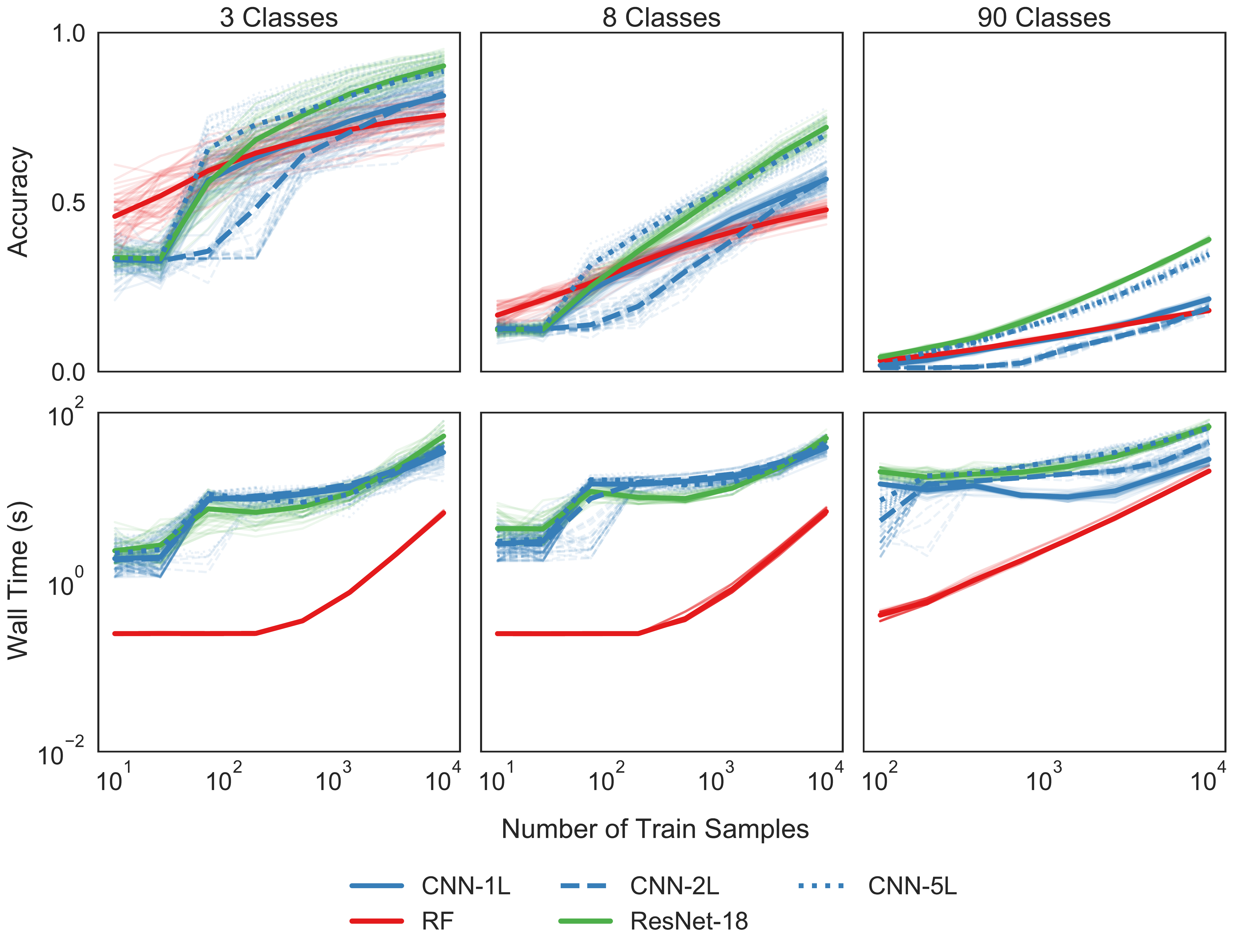}
  \caption{Performance of forests and networks on multiclass CIFAR-10/100 classifications with unbounded time and cost.
  Upper row shows classifier accuracy on a linear scale, and bottom row shows training wall times in seconds on a logarithmic scale. The x-axes correspond to logarithmic sample sizes for respective columns. Each panel shows average results over 45 random combinations. The left two columns use CIFAR-10, while the rightmost uses CIFAR-100.
  Compared to CNNs, RF performs better at small sample sizes and is always more efficient. As the class number increases, more complex networks have better classification accuracy and longer training wall times.
  }
\label{fig:cifar}
\end{figure}

\paragraph{Evaluation Criteria}
We evaluated the performance by classification accuracy and training wall times. The training wall times calculated the fitting time for the given model after hyperparameter tuning, measured in seconds. The provided test sets were used.

\subsubsection{Results}

\paragraph{Unbounded Time and Cost}
RF outperforms the networks at smaller sample sizes (Figure \ref{fig:cifar}). However, CNN accuracy often overcomes that of the classical models eventually. Higher class numbers decrease the accuracy of all models implemented, but the advantage of RF at smaller sample sizes also diminishes as the number of classes increases. ResNet-18 with early stops completely surpasses both classical models in the 90-class classification task. Among the networks, more convolutional layers produce higher accuracy, and the performance of 5-layer CNN is very close to that of ResNet-18. ResNet-18 always surpasses other models at 10,000 samples.

The training wall times of parallelized RF stay relatively constant until acquiring faster growths at larger sample sizes (Figure \ref{fig:cifar}). With early stopping, CNNs would produce training time descents at around 100 samples and increase along with sample size again with slower growth rates. The training time trajectories of CNNs partially overlap each other and always stay higher than those of RF. Only the 90-class task produces noticeable differences for CNNs' training wall times, which become more visible at larger sample sizes.


\begin{figure}[htb]
\centering
\includegraphics[width=0.8\textwidth]{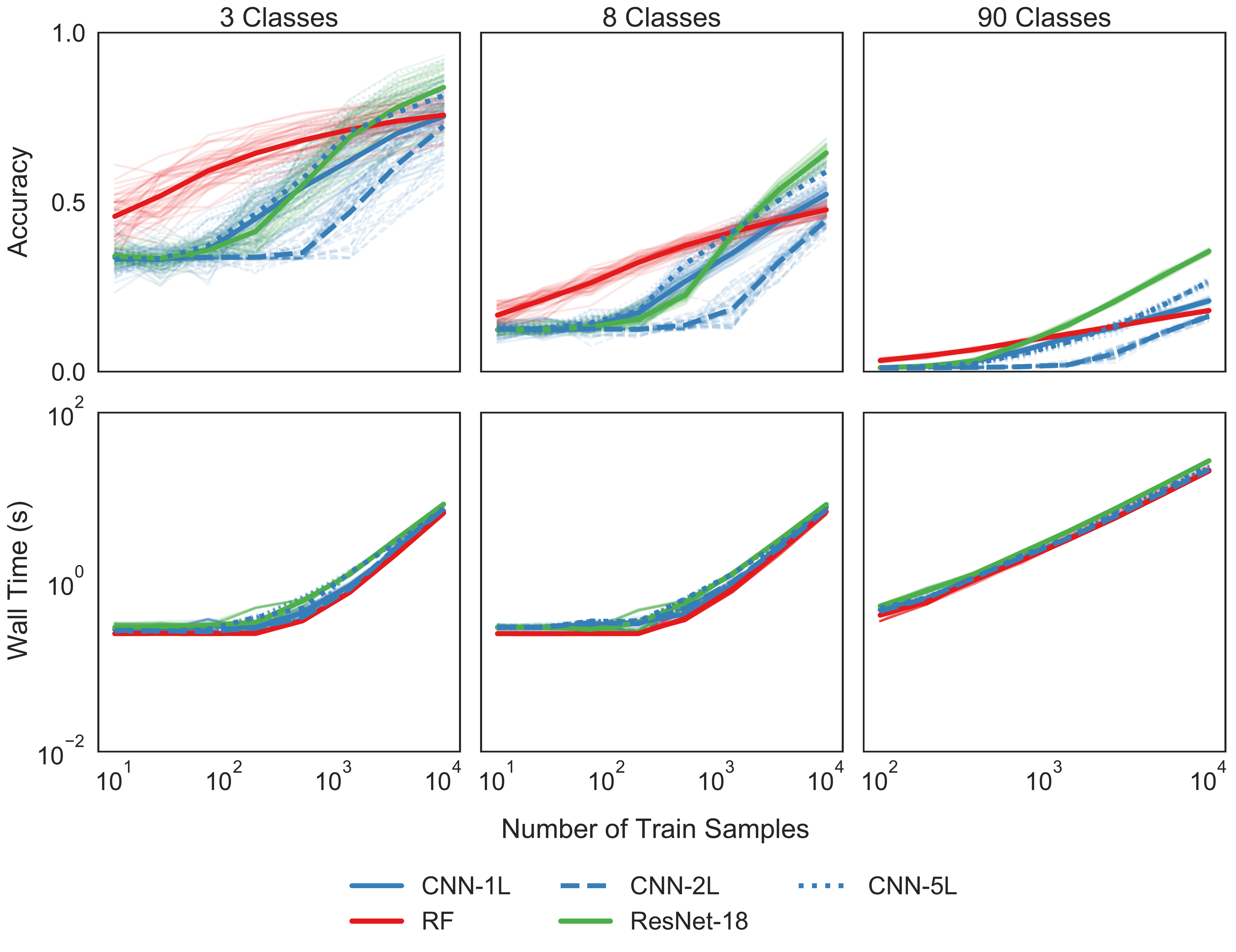}
  \caption{Performance of forests and networks on multiclass CIFAR-10/100 classifications with fixed training time.
  Upper row shows classifier accuracy on a linear scale, and bottom row shows training wall times in seconds on a logarithmic scale. The x-axes correspond to logarithmic sample sizes for respective columns. Each panel shows average results over 45 random combinations. The left two columns use CIFAR-10, while the rightmost uses CIFAR-100.
  RF has higher classification accuracy when compared to CNNs at smaller sample sizes. Complex networks, however, surpass RF at larger sample sizes, and ResNet-18 always performs best in the end.
  }
\label{fig:cifar_st}
\end{figure}

\paragraph{Fixed Training Time}
We then compared methods such that each took about the same time on the same virtual machine for 10,000 training samples. The baseline is RF's training times when run on the 6-core Standard\_NC6 Azure compute (Table \ref{table:azure}).
All training wall times conform to one shape (Figure \ref{fig:cifar_st}). Under the resource constraints, all the accuracy trajectories of CNNs are lower than those in the unbounded benchmarks (Figure \ref{fig:cifar}). Only at around 1,000 samples does CNN accuracy surpass that of RF, whereas 100 samples is the dividing point for benchmarks without restraints. ResNet-18 still has the highest accuracy at 10,000 samples, surpassing all other classifiers eventually. The results with fixed training cost are qualitatively similar (Figure \ref{fig:cifar_sc}).

Vision benchmarks with the CIFAR datasets show that networks would benefit from larger sample sizes and higher class numbers. More complex networks like ResNet-18 would achieve better performance (Figure \ref{fig:cifar}). In contrast, RF maintains the advantage on classification accuracy at small sample sizes, especially when the training times are fixed (Figure \ref{fig:cifar_st}).

\subsection{Audition}
\label{audition}
We performed benchmarks on the Free Spoken Digit Dataset (FSDD) (CC BY-SA 4.0) \citep{FSDD}. The dataset includes six different speakers pronouncing each of the 10 digits 50 times, for a total of 3,000 auditory recordings. Similar to the vision dataset analysis, we considered RF and various CNN architectures with different layers.

\subsubsection{Methods}
\paragraph{Computing}
For the 3-class and 8-class training sets, we sampled 480 recordings by selecting 45 random combinations of class labels and stratifying data equally among the classes. We also randomly selected 10\% of all the recordings for validation and another 10\% for benchmarks \citep{nasr, tian}.

To preprocess the auditory files for networks, we used the short-time Fourier transform to convert the 8 kHz raw auditory signals into spectrograms \citep{wyse}. In addition, we extracted mel-spectrograms (Appendix \ref{app:mel}) and mel-frequency cepstral coefficients (MFCCs) (Appendix \ref{app:mfcc}) using PyTorch's inbuilt functions \citep{pytorch}.
The size of fast Fourier transforms was set to 128 for spectrograms and mel-spectrograms, and 128 coefficients were retained for MFCC. We then scaled the data to zero mean and unit variance and reshaped the results into 32x32 single-channel images \citep{lecun2012efficient}. 

For RF, we still used \texttt{RandomForestClassifier} from the scikit-learn package \citep{scikit-learn}. Compared to the network architectures for vision data (Section \ref{vision}), the only modification we made was setting the channel input to one for all the simpler CNNs. However, the pre-trained ResNet-18 requires three channels for RGB colors. To accommodate this requirement, we concatenated the images to two duplicates of themselves along the channel dimension. This new approach would be called ResNet-18-Audio for differentiation.

All hyperparameters were the same as those of vision analysis. We benchmarked the models on the same Microsoft Azure compute with a 6-core CPU and a 1-core GPU: Standard\_NC6 (Table \ref{table:azure}).

\paragraph{Evaluation Criteria}
We evaluated the performance by classification accuracy. After training the classifiers, we benchmarked them using 10\% of the dataset that was left aside, giving us 30 auditory samples per class.
Thus, the test sets for the 3-class classification task have 90 auditory samples, while the 8-class test sets have 240 auditory samples.

\subsubsection{Results}

\begin{figure}[htb]
\centering
\includegraphics[width=0.6\textwidth]{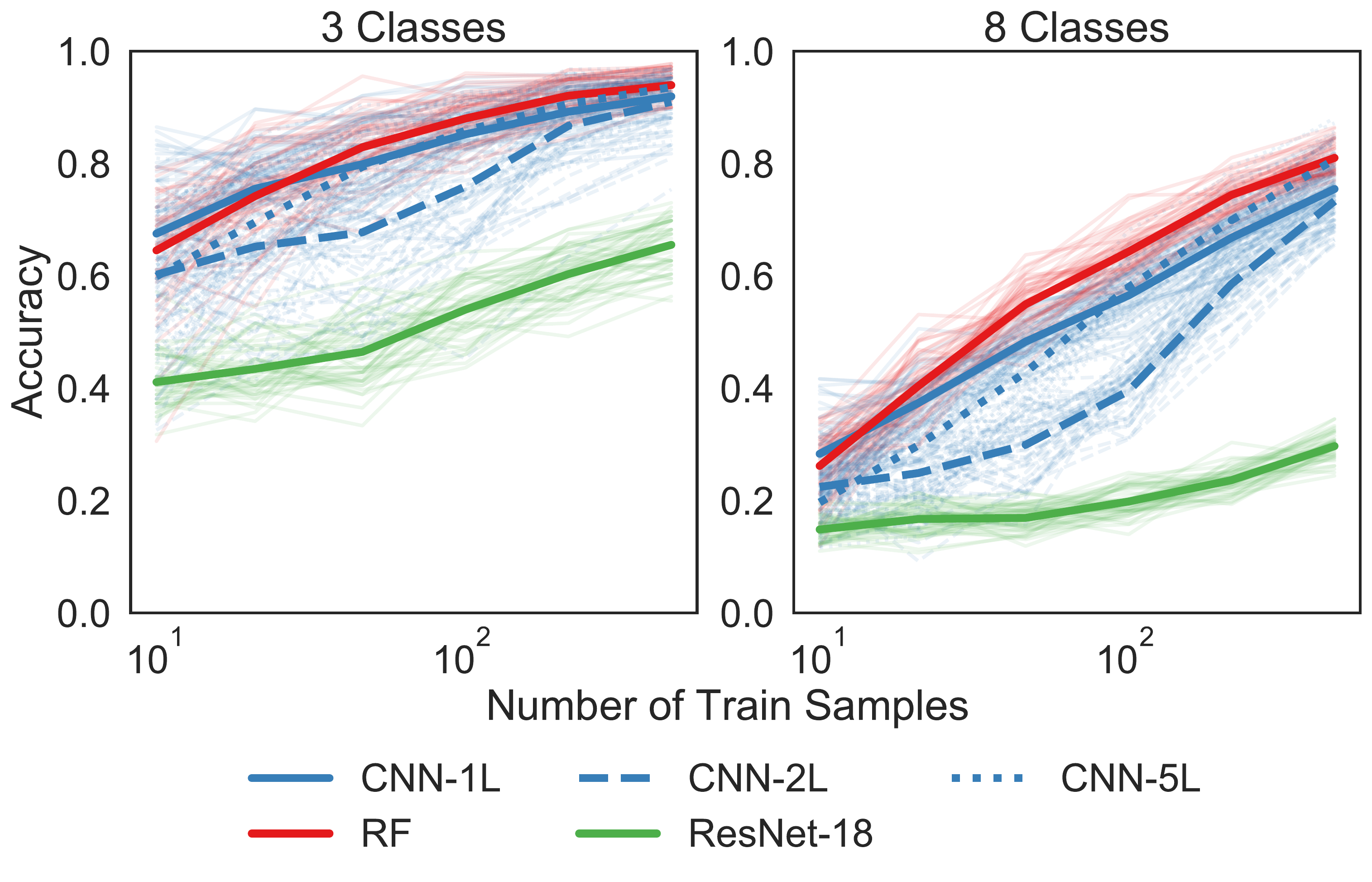}
  \caption{Performance of forests and networks on multiclass FSDD classifications using spectrograms. 
  The y-axes represent classifier accuracy on a linear scale and the x-axes correspond to logarithmic sample sizes from 10 to 480. Each panel shows average results over 45 random class combinations and individual trajectories with lower alpha.
  In the 3-class task, RF, 1-layer, and 5-layer CNNs all have very similar performances. In the 8-class task, RF achieves the highest accuracy. ResNet-18-Audio performs much worse than other classifiers in both tasks.
  }
\label{fig:spoken_digit}
\end{figure}

\paragraph{Spectrograms}
RF performs the best among all classifiers for essentially all sample sizes and all numbers of classes (Figure \ref{fig:spoken_digit}). ResNet-18-Audio has the worst performance in both tasks, presumably due to the small sample sizes that do not allow the complex network to train all its parameters effectively.
The accuracy in the 8-class task compared to 3-class is lower for all models across all sample sizes. Thus, forests excel at auditory classifications at small sample sizes, while simpler networks would perform better than complex ones.
The results with mel-spectrograms (Figure \ref{fig:mel}) and MFCCs (Figure \ref{fig:mfcc}) are qualitatively similar.

\section{Discussion}
\label{Discussion}
Conceptually, we described state-of-the-art machine learning methods simplified in classical statistical pattern recognition terms. The depiction of forests and networks as ``partition and vote'' schemes allows both a unified basic understanding of how these methods work from the perspective of classical classification and useful basic insight into their relationship with each other and potentially brain functions. 
Learning in biological brains can be viewed similarly as ensemble ``partition and vote'' functions implemented by a network of nodes.
In brains, a ``node'' can correspond to a unit across many scales, ranging from synaptic channels (which can be selectively activated or deactivated due to the synapses' local history), to cellular compartments, individual cells, or cellular ensembles \citep{Vogelstein2019-om}.
At each scale, brains learn by partitioning feature space that is the set of all possible sensory inputs; a ``part'' corresponds to a subset of ``nodes'' that tend to respond to any given input.
An example is the selective response properties that define cortical columns---columns in the sensory cortex \citep{Mountcastle1997-by}.
Brains also vote, where voting is a pattern of responses based on neural activation that indicate which stimulus evoked a response \citep{Machens2005-wk}.
See Appendix \ref{app:brains} for further details.

Empirically, we provided comparisons for forests and networks on three data modalities and produced consistent results. Importantly, in the structured data scenarios, the input to networks typically includes not just the magnitude of each feature, but also the relative position of the features (for example, image pixels comprise a local patch as encoded into a convolutional layer). This is in contrast to forests, for which the input is purely the magnitude, without the relative location, of each feature. Thus, if each feature is encoded by a triple $(m,x,y)$, including the magnitude, horizontal, and vertical positions of the feature, forests tend to only get $1/3$ of the input that the networks get, significantly impoverishing the information they have to start with. In general, we found forests to excel at tabular and structured data (vision and audition) at small sample sizes, whereas networks performed better on structured data at larger sample sizes. 

There do exist some limitations to this technical report, which we are planning to address in the coming months. Our next steps include: adding more metrics and stratifications to results, optimizing hyperparameter search for each sample size, and including other estimators in benchmarks, such as GBDT and sparse projection oblique randomer forests (SPORF) \citep{sporf}. Benchmark code is publicly accessible at our website: \url{https://dfdn.neurodata.io/}.


\bibliographystyle{unsrtnat}
\bibliography{ref}

\clearpage

\appendix

\section{CIFAR-10/100 Benchmarks with Fixed Training Cost}
\label{app:cifar_sc}
\begin{figure}[!htb]
\centering
\includegraphics[width=0.8\textwidth]{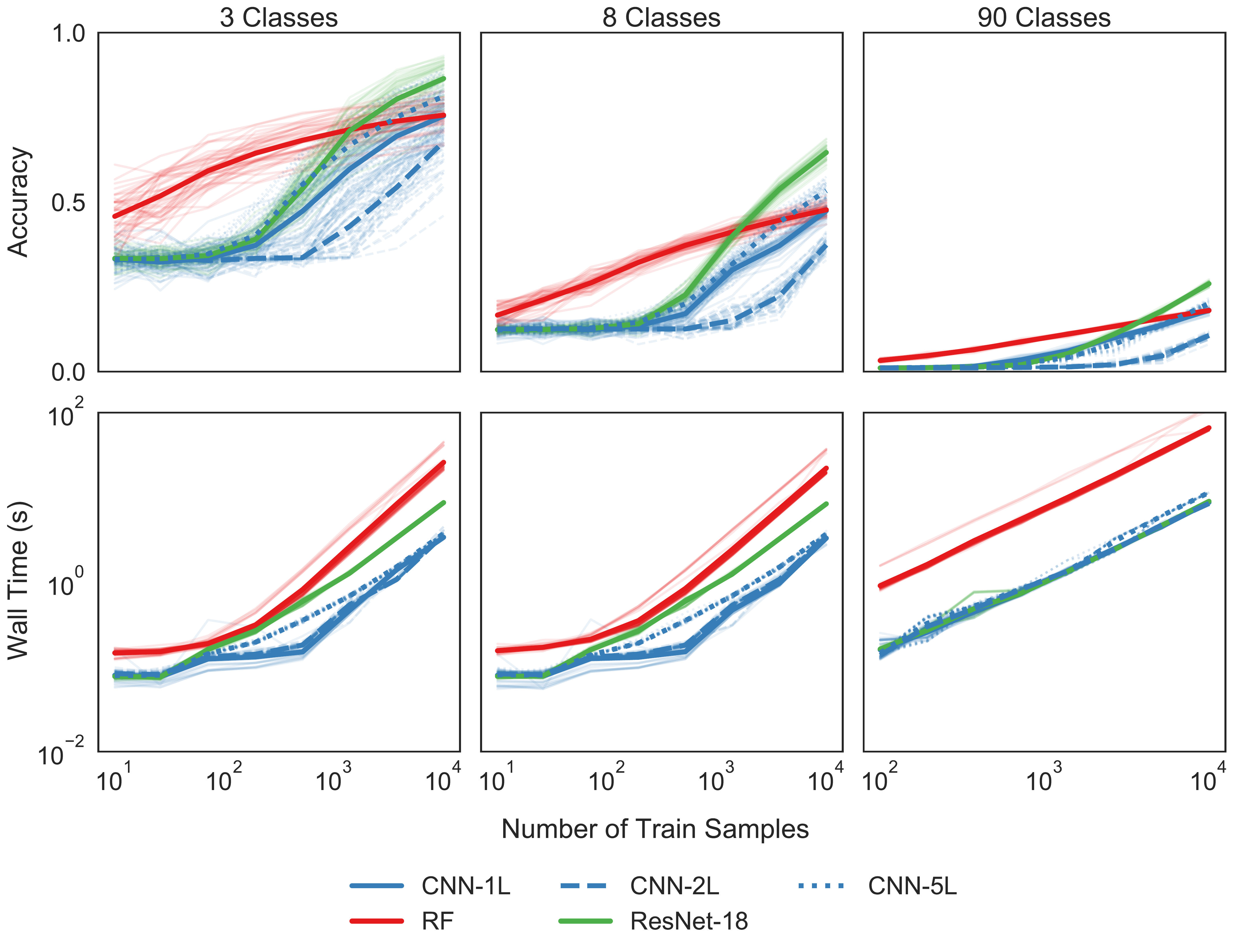}
  \caption{Performance of forests and networks on multiclass CIFAR-10/100 classifications with fixed training cost.
  Upper row shows classifier accuracy on a linear scale, and bottom row shows training wall times in seconds on a logarithmic scale. The x-axes correspond to logarithmic sample sizes for respective columns. Each panel shows average results over 45 random combinations. The left two columns use CIFAR-10, while the rightmost uses CIFAR-100.
  RF has higher classification accuracy when compared to CNNs at smaller sample sizes. Complex networks, however, surpass RF at larger sample sizes, and ResNet-18 always performs best in the end.
  }
\label{fig:cifar_sc}
\end{figure}

We also compared methods such that each took about the same cost on two virtual machines for 10,000 training samples (Figure \ref{fig:cifar_sc}). The baseline is RF's training times as run on the 2-core Standard\_DS2\_v2 Azure compute instance (Table \ref{table:azure}). 
As a result, the training wall times of CNNs, which often use the minimum epoch number, are always lower than those of RF. Due to the CNNs' different complexities, the correspondence between training costs becomes more accurate as the class number increases. The networks' training time trajectories also overlap more completely. The results are qualitatively similar to CIFAR benchmarks with fixed training time (Figure \ref{fig:cifar_st}).

\clearpage

\section{SVHN Benchmarks}
\label{app:svhn}
\begin{figure}[!htb]
\centering
\includegraphics[width=0.6\textwidth]{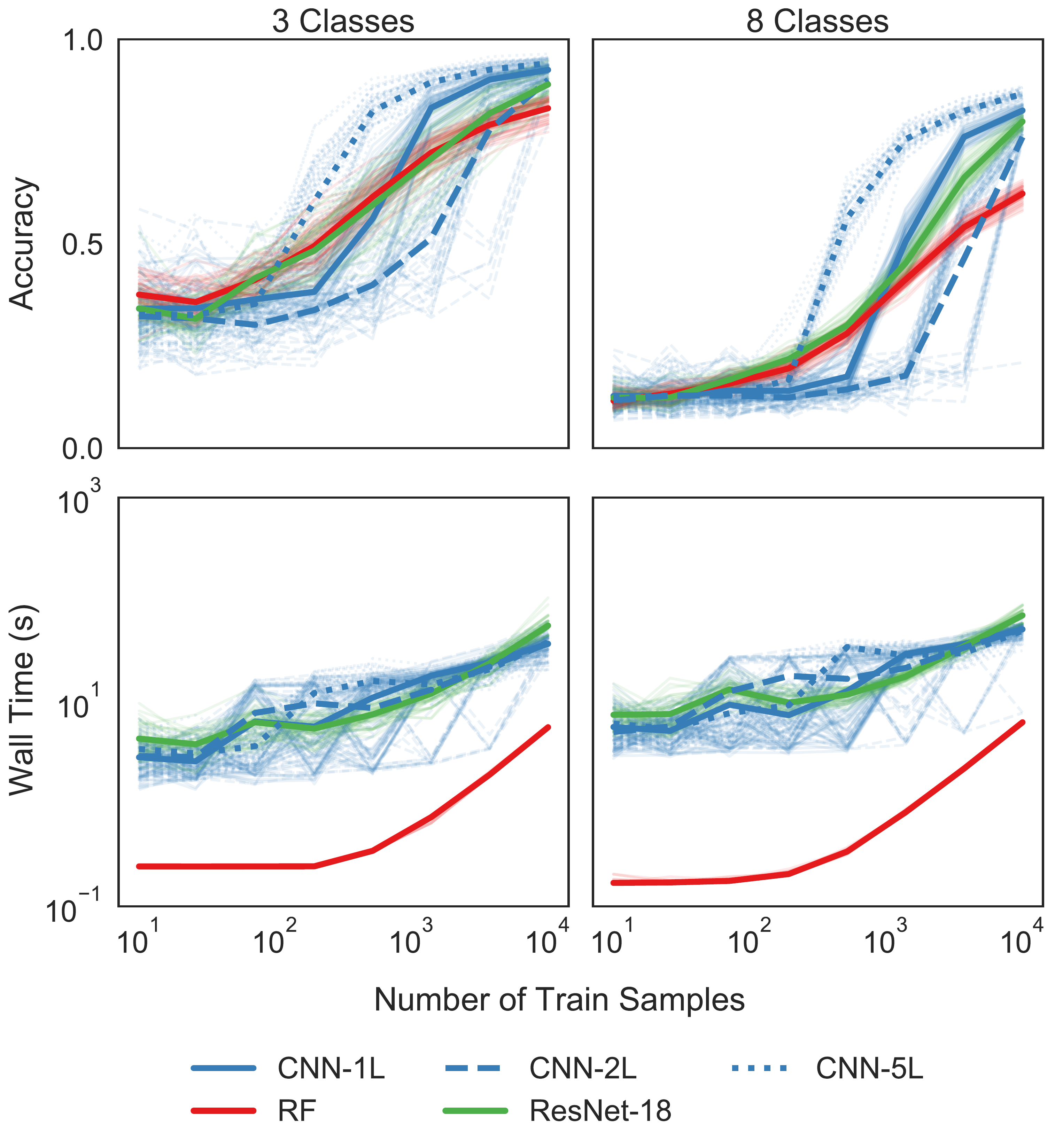}
  \caption{Performance of forests and networks on multiclass SVHN classifications with unbounded time and cost.
  Upper row shows classifier accuracy on a linear scale, and bottom row shows training wall times in seconds on a logarithmic scale. The x-axes correspond to logarithmic sample sizes for respective columns. Each column shows average results over 45 random combinations.
  Compared to CNNs, RF performs better and faster at smaller sample sizes.
  }
\label{fig:svhn}
\end{figure}

The SVHN dataset contains 73,257 digits for training and 26,032 for testing \citep{svhn}. The 3-class and 8-class tasks show surprising trends for networks, as simpler CNNs surpass ResNet-18 on classification accuracy as sample size increases. At higher sample sizes, 5-layer CNN has the best performance among all classifiers. Network accuracy is always higher than that of RF at 10,000 samples (Figure \ref{fig:svhn}). Although RF performs better than networks at smaller sample sizes in the 3-class task, the advantages disappear in the 8-class task. As seen in the CIFAR benchmarks (Figure \ref{fig:cifar}, \ref{fig:cifar_st}, \ref{fig:cifar_sc}), networks would be more adept at handling higher class numbers.

The trends of training wall times are very similar to those of CIFAR benchmarks with unbounded time and cost (Figure \ref{fig:cifar}). Forests' training times are always shorter than networks', and more fluctuations occur for CNN trajectories.

\clearpage

\section{FSDD Benchmarks with Mel-Spectrograms}
\label{app:mel}
\begin{figure}[!htb]
\centering
\includegraphics[width=0.6\textwidth]{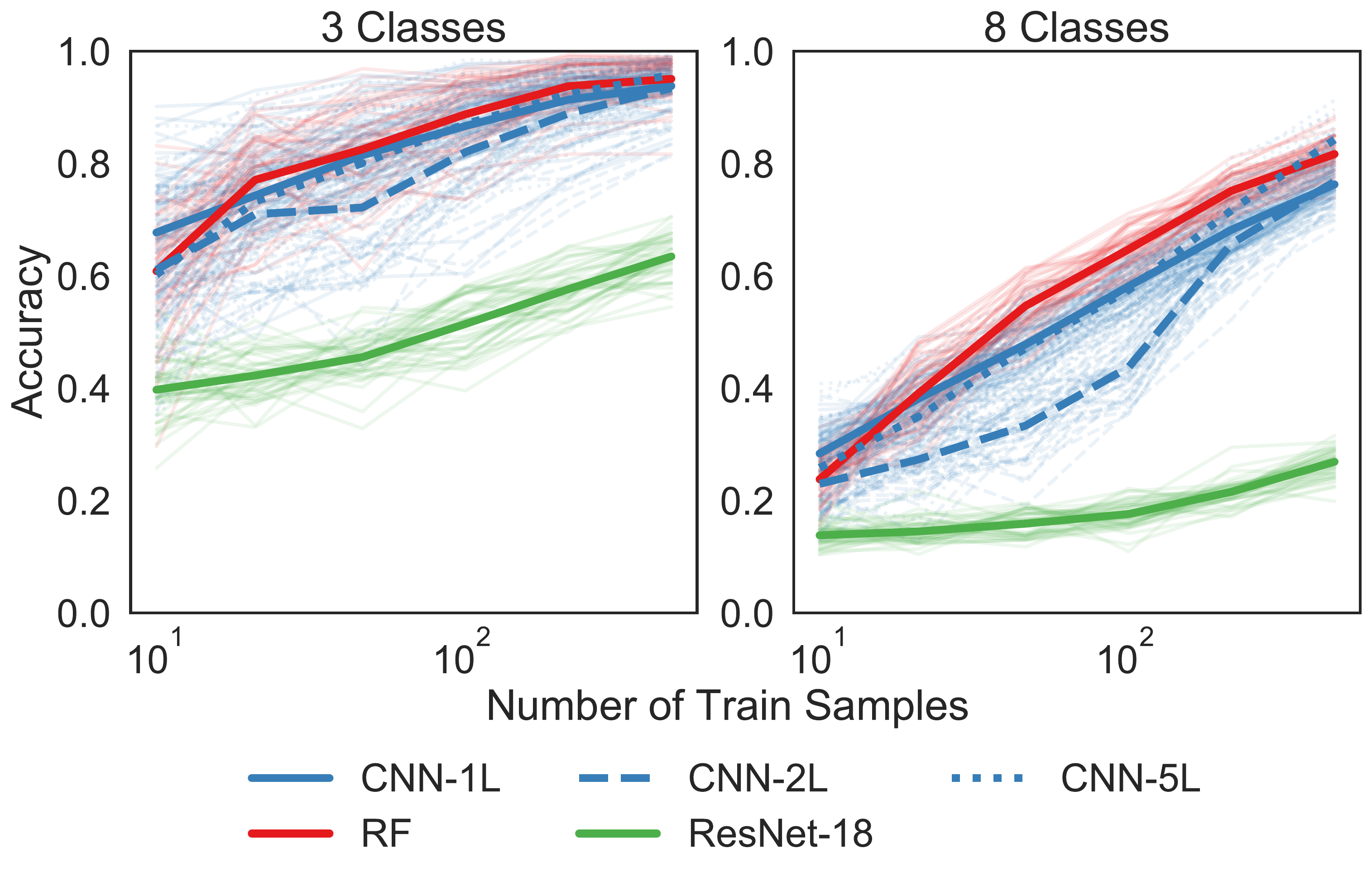}
  \caption{Performance of forests and networks on multiclass FSDD classifications using mel-spectrograms.
  The y-axes represent classifier accuracy on a linear scale and the x-axes correspond to logarithmic sample sizes from 10 to 480. Each panel shows average results over 45 random class combinations and individual trajectories with lower alpha
  In the 3-class task, RF, 1-layer, and 5-layer CNNs all have very similar performances. In the 8-class task, RF achieves the highest accuracy. ResNet-18-Audio performs much worse than other classifiers.
  }
\label{fig:mel}
\end{figure}

As an alternative approach, we used PyTorch’s inbuilt function and converted the raw audio magnitudes into mel-spectrograms \citep{pytorch}. The process involves the aforementioned spectrogram conversions and uses triangular filterbanks to modify the images. The results (Figure \ref{fig:mel}) are qualitatively similar to FSDD benchmarks with spectrograms (Figure \ref{fig:spoken_digit}) and MFCCs (Figure \ref{fig:mfcc}).

\clearpage

\section{FSDD Benchmarks with MFCCs}
\label{app:mfcc}
\begin{figure}[!htb]
\centering
\includegraphics[width=0.6\textwidth]{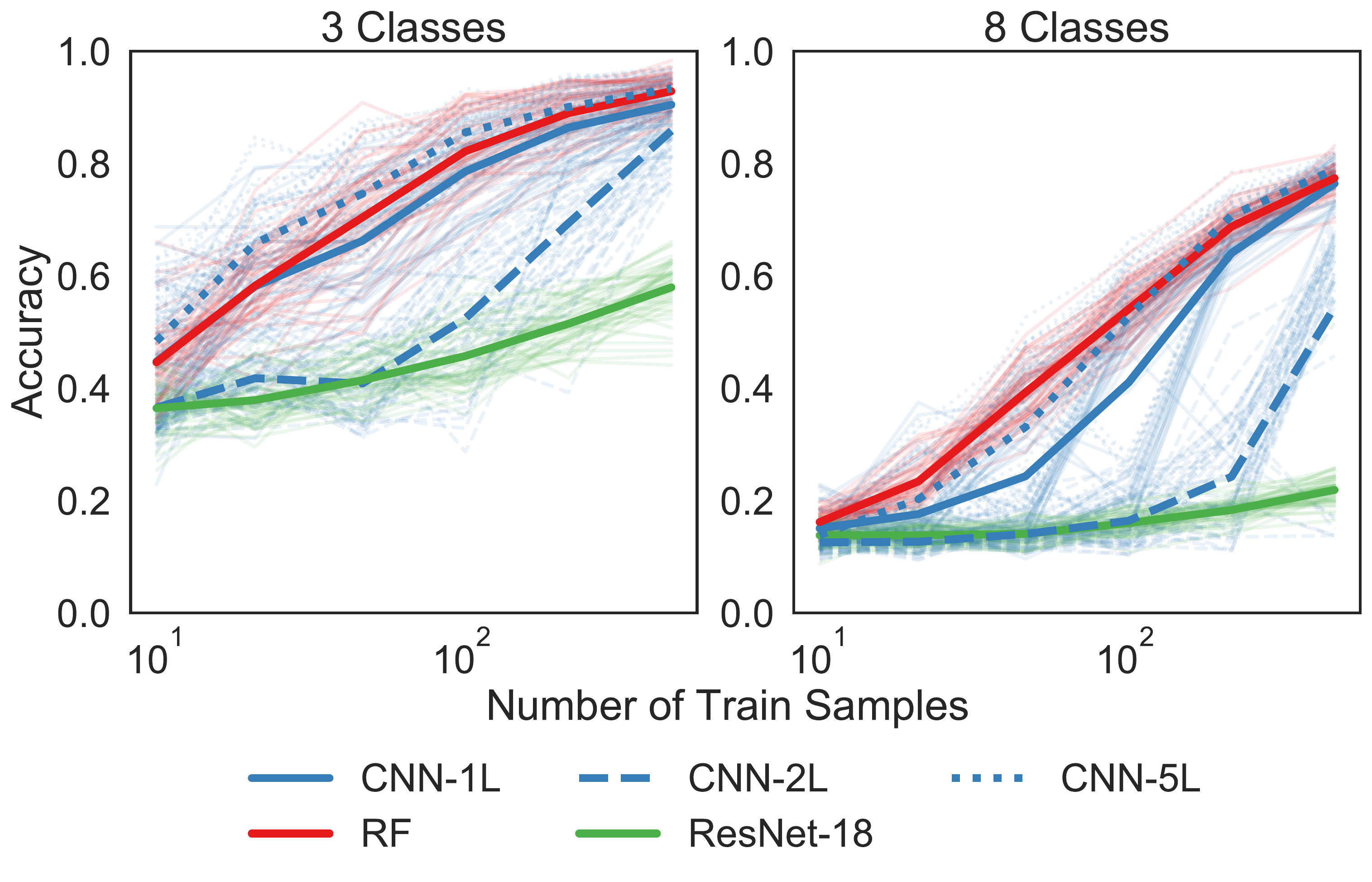}
  \caption{Performance of forests and networks on multiclass FSDD classifications using mel-frequency cepstral coefficients (MFCC). 
  The y-axes represent classifier accuracy on a linear scale and the x-axes correspond to logarithmic sample sizes from 10 to 480. Each panel shows average results over 45 random class combinations and individual trajectories with lower alpha.
  At the maximum sample size, RF, 1-layer, and 5-layer CNNs all have very similar accuracy. 2-layer CNN performs worse, while ResNet-18-Audio performs the worst.
  }
\label{fig:mfcc}
\end{figure}

As an alternative conversion, we used PyTorch’s inbuilt function and converted the raw audio magnitudes into MFCC \citep{pytorch}. The process calculates MFCCs on the DB-scaled mel-spectrograms. The results (Figure \ref{fig:mfcc}) are qualitatively similar to FSDD benchmarks with spectrograms (Figure \ref{fig:spoken_digit}) and mel-spectrograms (Figure \ref{fig:mel}).

\clearpage

\section{Zebrafish Brain Maps}
\label{app:brains}
\begin{figure}[h!]
\centering
\includegraphics[width=0.7\textwidth]{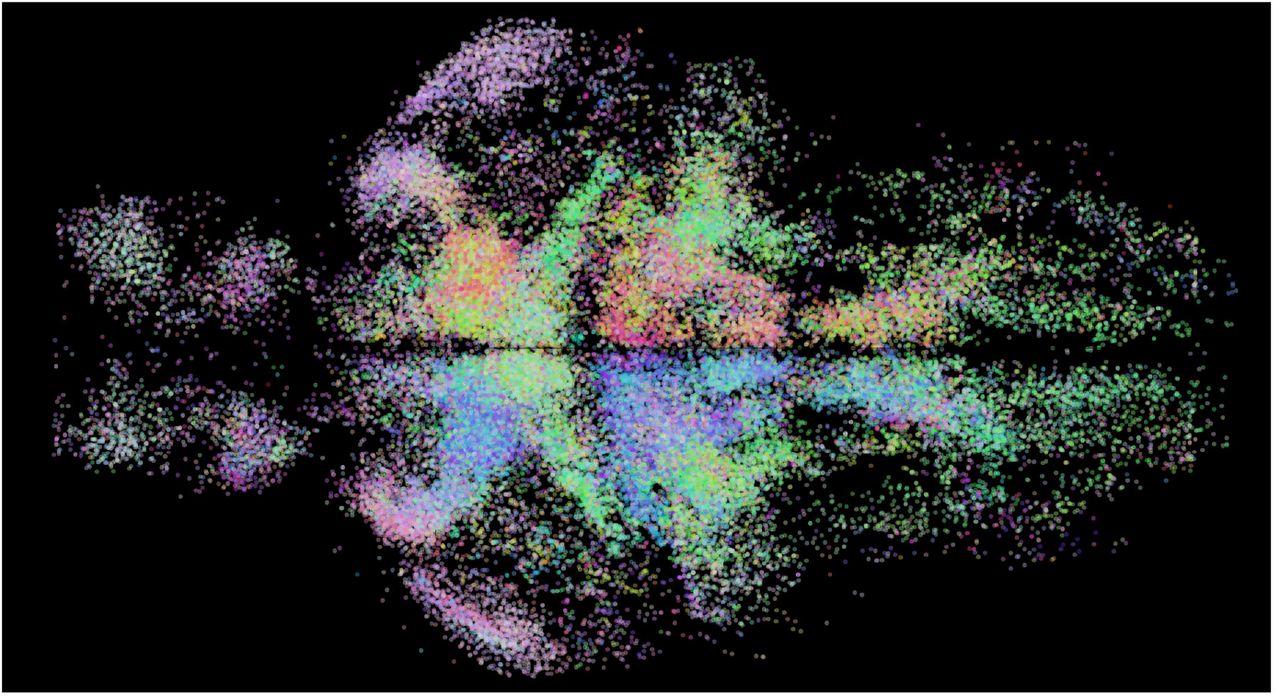}
  \caption{Whole-brain activity maps reveal processing stages underlying the optomotor response in a larval zebrafish.
  }
\label{fig:fluorian}
\end{figure}

Figure \ref{fig:fluorian} illustrates neural selectivity to features of sensory input in a larval zebrafish, and indicates a learned partition and vote framework for brain functioning \citep{Naumann2016-oc}.
Specifically, this image shows all motion-sensitive nodes ($n = 76,604$) in a brain. Each node is a dot, color coded for preferred motion direction.
The image indicates that each node's activity indeed corresponds to a partition of feature space, which is actively refined through sensory experience and learning.
A relationship with forests and networks, at least at a basic level, is clear.
The utility of this relationship, for either machine learning or neuroscience, is the subject of endless conjecture and refutation.

\clearpage

\end{document}